\documentclass[letterpaper, 10 pt, conference]{ieeeconf}  %

\IEEEoverridecommandlockouts                              %

\usepackage{multirow}
\usepackage{booktabs}
\usepackage{siunitx}
\usepackage{tabularx}
\newcolumntype{C}{>{\centering\arraybackslash}X}
\usepackage[skip=0.333\baselineskip]{caption}
\usepackage{booktabs,siunitx}
\usepackage{newtxtext,newtxmath} %

\usepackage{subcaption}

\usepackage{esvect}

\usepackage{amsmath}
\usepackage{xcolor} 

\usepackage{mathtools}
\newcommand{\defeq}{\vcentcolon=}

\usepackage{hyperref}

\usepackage{placeins}

\usepackage{mathtools}
\DeclarePairedDelimiterX{\infdivx}[2]{(}{)}{%
  #1\;\delimsize\ \mid \;#2%
}

\title{\LARGE \bf
Learning Distributional Demonstration Spaces for Task-Specific Cross-Pose Estimation}

\author{Jenny Wang$^{1}\textsuperscript{*}$, Octavian Donca$^{1}\textsuperscript{*}$, and David Held$^{1}$%
\thanks{This material is based upon work supported by the United States United States Air Force and DARPA under Contract No. FA8750-18-C-0092, NIST under Grant No. 70NANB23H178, and the Uber Presidential Fellowship.}%
\thanks{$^{1}$All authors are with the Robotics Institute, Carnegie Mellon University. $\textsuperscript{*}$ represents equal contribution. (\href{mailto:jennyw2@andrew.cmu.edu}{\texttt{jennyw2@andrew.cmu.edu} },
\href{mailto:odonca@andrew.cmu.edu}{\texttt{odonca@andrew.cmu.edu} }, \href{mailto:dheld@andrew.cmu.edu}{\texttt{dheld@andrew.cmu.edu} })}%
}

\begin{document}

\maketitle
\thispagestyle{empty}
\pagestyle{empty}

\begin{abstract}
    Relative placement tasks are an important category of tasks in which one object needs to be placed in a desired pose relative to another object.  Previous work has shown success in learning relative placement tasks from just a small number of demonstrations when using relational reasoning networks with geometric inductive biases. However, such methods cannot flexibly represent multimodal tasks, like a mug hanging on any of $n$ racks. We propose a method that incorporates additional properties that enable learning multimodal relative placement solutions, while retaining the provably translation-invariant and relational properties of prior work. We show that our method is able to learn precise relative placement tasks with only 10-20 multimodal demonstrations with no human annotations across a diverse set of objects within a category. Supplementary information can be found on the website: \href{https://sites.google.com/view/tax-posed/home}{\color{blue}{https://sites.google.com/view/tax-posed/home}}.
\end{abstract}

\section{Introduction}
\label{sec:introduction}

Many robotic manipulation tasks can be framed as relative placement tasks. For example, hanging a mug on a mug rack requires placing the mug in a position relative to one of the pegs of the rack. %
Even complex, long-horizon tasks such as organizing a cluttered table can be framed as a series of relative placements: first, predict an SE(3) transformation that stacks one book on top of another, then predict a transformation that puts the pencil in the pencil box, and then predict a transformation that centers the keyboard in front of the monitor. %

Previous work such as TAX-Pose~\cite{pan2023tax} has shown that, for relative placement tasks, using network architectures that explicitly reason about object relationships helps the network to generalize significantly better across object poses and instances. However, this previous work outputs only a single relative placement prediction for each observation. %
In multi-modal settings, this  leads to predictions which are the mean of valid placement modes, 
which may be incorrect. For example, suppose that a set of demonstrations place a mug on any of $n$ racks in the scene. The average of these demonstrations will be a point in the middle of the racks, which is not a valid placement.
In contrast,  many tasks are defined by a distribution of relationships: a robot may be tasked to grasp anywhere along the rim of a bowl, place a fork on the left of any of the plates  (e.g. when setting the table), or grasp any one of a cabinet's drawers.

To address these challenges, we present TAX-PoseD, a Distributional 
variant of TAX-Pose~\cite{pan2023tax}. Our method predicts task-specific object relationships from just a few demonstrations and no human annotations, while robustly accounting for multimodal demonstration distributions. %

Our core technical contributions include:

\begin{itemize}
    \item A method for efficiently learning distributional relative placement tasks; our approach extends  TAX-Pose~\cite{pan2023tax} to handle multimodal, distributional demonstrations. %
    \item A novel spatially-grounded architecture for a cVAE~\cite{sohn2015learning}  that represents the latent variable distribution as a categorical distribution over 3D points; this results in a grounded and interpretable latent space that avoids the smoothing effect commonly found in cVAEs, leading to significantly improved performance for multi-modal placement tasks.%
\end{itemize}

We evaluate our method on challenging multimodal tasks and evaluate its generalization across diverse objects within a category. We demonstrate that our method is both interpretable and achieves strong performance on distributional relative placement tasks. 

\begin{figure}[t!]
    \centering
    \vspace{0.2cm} %
    \includegraphics[width=0.48\textwidth]{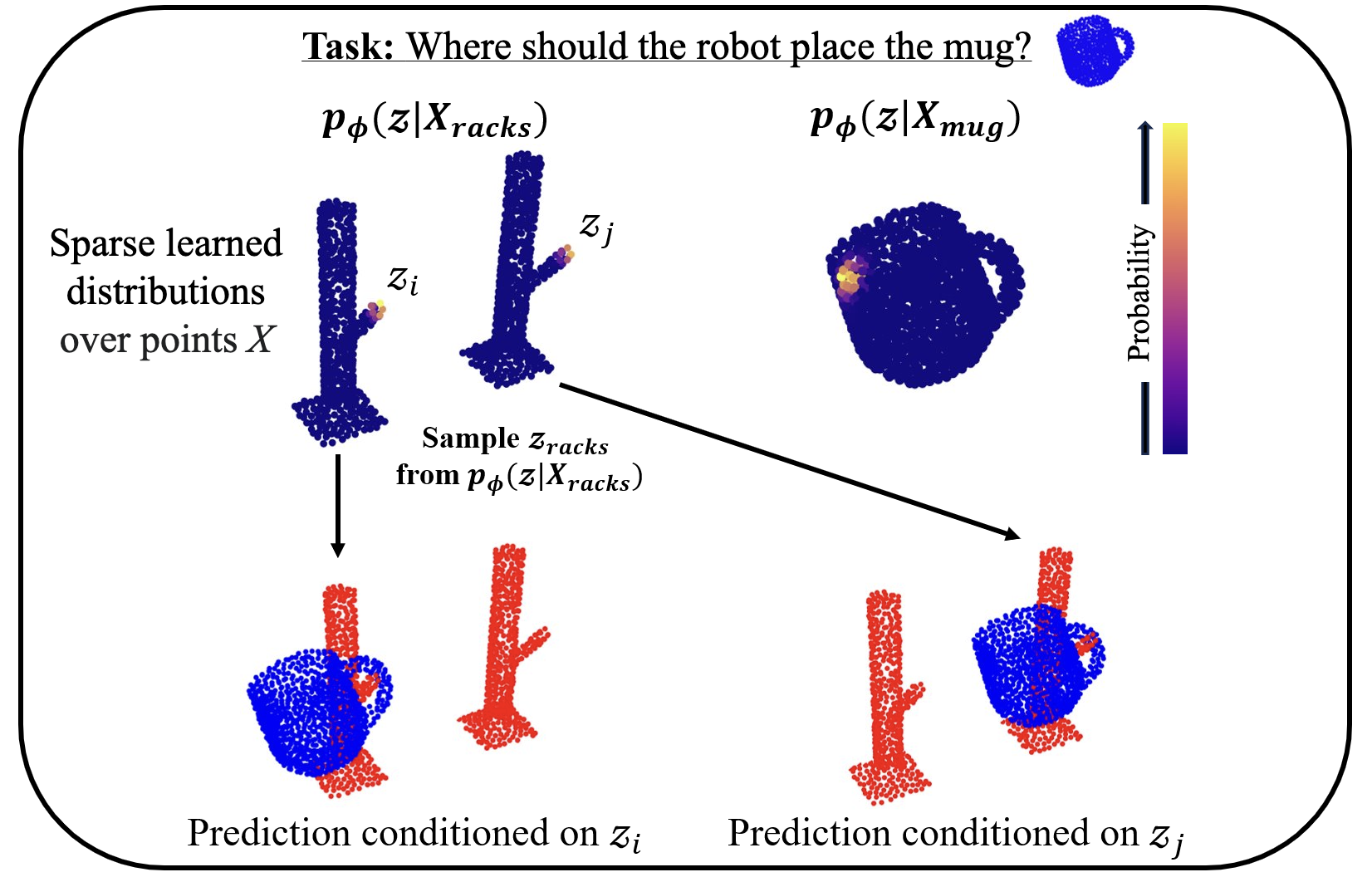}
    \caption{Our method's  learned prior $p_\phi(z \mid X)$ learns a distribution over modalities for the task, for example placing a mug on the left rack or the right rack. During inference time, this allows the model to predict a diverse set of ways to perform the task. }%
    \label{fig:p(z|X)_vis.jpg}
    \vspace{-0.5cm}
\end{figure}

\section{Related Work}
\label{sec:related_work}

\textbf{Action Representations.} Many action representations have been explored that enable robots to increase their learning efficiency when learning to solve manipulation tasks. For example, representations can be per-point~\cite{wu2021vat,mo2021where2act,eisner2022flowbot3d} or consist of keypoints~\cite{qin2020keto}. %
Additionally, architectures that leverage local geometry have been shown to provide useful priors for learning from point cloud data~\cite{kipf2016semi,wang2019dynamic,qi2017pointnet++,zhao2021point}. %
Our proposed method combines the strengths of dense per-point representations and geometric inductive biases while also reasoning over distributions for relative placement tasks. %

\textbf{Relative Placement Tasks:}
Many tasks can be defined as a sequence of relative placement tasks, such as hanging a mug on a rack or  putting a dish into a microwave. These tasks are ``relative" placements because the desired pose of the first object is not based on a fixed set of coordinates in the world frame but rather it is defined relative to some other object in the scene. A number of papers have developed methods designed for this class of tasks~\cite{simeonov2022neural, pan2023tax, simeonov2023se, simeonov2023shelving}.  This paper builds upon TAX-Pose~\cite{pan2023tax}, which achieves strong performance on unimodal tasks such as hanging a single mug on a single rack; however, as we show in this work, TAX-Pose performs poorly on multimodal tasks such as when there are multiple mug racks available. Our method handles such symmetries by learning a distribution over relative placements. Symmetries are also addressed by concurrent work~\cite{simeonov2023shelving}, which uses diffusion models.  However, diffusion models are often slow during inference; in contrast, our method can operate at high inference speeds because we require only a single forward pass through our network.

\textbf{Multimodal Action Sets.} Actions that successfully complete a task may occupy multiple regions of an agent's action space. When learning from demonstrations, an architecture that does not model the multimodality of the demonstration space may tend to output the average over correct actions, which leads to failures. Implicit representations have been explored for modeling the multimodality of a task~\cite{florence2022implicit}. 
Multi-modal action reasoning can be implemented as spatial action maps~\cite{wu2020spatial}, GANs~\cite{pandeva2019mmgan}, particle filters~\cite{wirnshofer2020controlling}, diffusion models~\cite{chi2023diffusion,simeonov2023shelving}, and VAEs~\cite{allshire2021laser, mousavian20196}.  We build upon the VAE literature and apply it to modeling modes of the demonstration space for relative placement tasks.

\section{Problem Statement and Assumptions}
\label{sec:problem_statement}

We consider the task of relative placement, 
for which an object must be placed in some configuration relative to another object.  For example, for the mug hanging task, a mug must be placed in some pose relative to a mug rack.  

Following Pan et al~\cite{pan2023tax}, success for a relative placement task can be defined as follows: Let $\textbf{\text{T}}^{*}_{\mathcal{A}}$ and $\textbf{\text{T}}^{*}_{\mathcal{B}}$ be a pair of SE(3) poses for objects $\mathcal{A}$ and $\mathcal{B}$, respectively (in some fixed world frame), for which the task is considered successful. For example, $\textbf{\text{T}}^{*}_{\mathcal{A}}$ can be the pose for a mug, and $\textbf{\text{T}}^{*}_{\mathcal{B}}$ can be the pose for a mug rack, for which the mug is on the mug rack. Given a pair of poses $\textbf{\text{T}}^{*}_{\mathcal{A}}$ and $\textbf{\text{T}}^{*}_{\mathcal{B}}$ that complete the task, for relative placement tasks, the task is also considered successful when any SE(3) transformation $\textbf{\text{T}}$ is applied to both objects $\mathcal{A}$ and $\mathcal{B}$. In other words, for a relative placement task, if the task is successful for poses $\textbf{\text{T}}^{*}_{\mathcal{A}}$ and $\textbf{\text{T}}^{*}_{\mathcal{B}}$, then the task is also successful for poses $\textbf{\text{T}} \cdot \textbf{\text{T}}^{*}_{\mathcal{A}}$
 and $\textbf{\text{T}} \cdot \textbf{\text{T}}^{*}_{\mathcal{B}}$. 
 Following this reasoning, suppose that objects $\mathcal{A}$ and $\mathcal{B}$ are in poses $\textbf{\text{T}}_{\mathcal{A}}$ and $\textbf{\text{T}}_{\mathcal{B}}$, respectively, and suppose that $\textbf{\text{T}}^{*}_{\mathcal{A}}$ and $\textbf{\text{T}}^{*}_{\mathcal{B}}$ are defined as above.  %
 Then 
  poses $\textbf{\text{T}}_{\mathcal{A}}$ and $\textbf{\text{T}}_{\mathcal{B}}$ must satisfy the following Boolean function to be considered a successful relative placement:

\vspace{-0.5cm}

\begin{align}
    \label{eq:relplace}
    \text{RelPlace}(\textbf{\text{T}}_{\mathcal{A}}, \textbf{\text{T}}_{\mathcal{B}}) =&\textbf{ SUCCESS}  \\
    &\text{ iff } \exists\textbf{\text{T}} \in \text{ SE(3)} \nonumber \\
    & \text{ s.t. } \textbf{\text{T}}_{\mathcal{A}} = \textbf{\text{T}} \cdot \textbf{\text{T}}^{*}_{\mathcal{A}}\text{ and }\textbf{\text{T}}_{\mathcal{B}} = \textbf{\text{T}} \cdot \textbf{\text{T}}^{*}_{\mathcal{B}}.  \nonumber
    \vspace{-0.5cm}
\end{align}
In practice, we are usually not given the transforms $\textbf{\text{T}}_{\mathcal{A}}$ and $\textbf{\text{T}}_{\mathcal{B}}$ but rather we are given the point clouds for these objects $\textbf{\text{P}}_{\mathcal{A}}$ and $\textbf{\text{P}}_{\mathcal{B}}$.
In order to solve a relative placement task, we need to estimate an SE(3) transformation $\textbf{\text{T}}_{\mathcal{A}\mathcal{B}} \defeq f(\textbf{\text{P}}_{\mathcal{A}}, \textbf{\text{P}}_{\mathcal{B}})$ that transforms object $\mathcal{A}$ %
in such a manner that satisfies RelPlace (Eqn.~\ref{eq:relplace}). In other words, we need to find the transform $\textbf{\text{T}}_{\mathcal{A}\mathcal{B}}$ such that $\text{RelPlace}(\textbf{\text{T}}_{\mathcal{A}\mathcal{B}} \cdot 
\textbf{\text{T}}_{\mathcal{A}}, \textbf{\text{T}}_{\mathcal{B}}) =\textbf{ SUCCESS}$.
In TAX-Pose~\cite{pan2023tax}, this transformation $\textbf{\text{T}}_{\mathcal{A}\mathcal{B}}$ 
is referred to as the \textit{cross-pose} between objects $\mathcal{A}$ and $\mathcal{B}$.

Often, there is a set of valid solutions that all solve the task, for example when a plate should be placed within any of the placemats on a table. %
Thus, we extend the definition of task success given by Equation~\ref{eq:relplace} to the distributional setting. For distributional relative placement tasks, suppose that object $\mathcal{B}$ is in pose $\textbf{\text{T}}^*_{\mathcal{B}}$. Then let $\{\textbf{\text{T}}^{*}_{\mathcal{A},1}, \ldots, \textbf{\text{T}}^{*}_{\mathcal{A},N}\}$ be a set of $N$ poses for object $\mathcal{A}$ that satisfy the relative placement task. Because this is a relative placement task, then the task is also successful for poses $\textbf{\text{T}} \cdot \textbf{\text{T}}^{*}_{\mathcal{A},i}$
 and $\textbf{\text{T}} \cdot \textbf{\text{T}}^{*}_{\mathcal{B}}$ (for objects $\mathcal{A}$ and $\mathcal{B}$ respectively) for all $i \in [1, N]$ for any SE(3) transformation $\textbf{\text{T}}$.
Thus, suppose that objects $\mathcal{A}$ and $\mathcal{B}$ are in poses $\textbf{\text{T}}_{\mathcal{A}}$ and $\textbf{\text{T}}_{\mathcal{B}}$, respectively.  Then 
 these poses must satisfy the following Boolean function to be considered a successful relative placement for a \emph{distributional} relative placement task:
\begin{align}
    \text{RelPlace}_D(\textbf{\text{T}}_{\mathcal{A}}, \textbf{\text{T}}_{\mathcal{B}}) =&\textbf{ SUCCESS} \label{eq:relplaced} \\
    &\text{ iff } \exists i \in [1, N], \exists\textbf{\text{T}} \in \text{SE(3)} \nonumber \\
    &\text{ s.t. } \textbf{\text{T}}_{\mathcal{A}} = \textbf{\text{T}} \cdot \textbf{\text{T}}^{*}_{\mathcal{A},i}\text{ and }\textbf{\text{T}}_{\mathcal{B}} = \textbf{\text{T}} \cdot \textbf{\text{T}}^{*}_{\mathcal{B}}. \nonumber
\end{align}

In practice, we are not usually given a comprehensive set of poses $\{\textbf{\text{T}}^{*}_{\mathcal{A},1}, \ldots, \textbf{\text{T}}^{*}_{\mathcal{A},N}\}$ for object $\mathcal{A}$ that solve a relative placement task relative to some pose $\textbf{\text{T}}^*_{\mathcal{B}}$ for object $\mathcal{B}$. Instead, we are often given a set of demonstration pairs $(\textbf{\text{T}}^{*}_{\mathcal{A},j}, \textbf{\text{T}}^{*}_{\mathcal{B},j})$ for $j \in [1, M]$ for a set of M demonstrations. From these demonstration pairs, we aim to learn a generative function $f_D(\textbf{\text{P}}_{\mathcal{A}}, \textbf{\text{P}}_{\mathcal{B}})$ such that sampling from this function gives a solution to the distributional relative placement task, where $\textbf{\text{P}}_{\mathcal{A}}$ and $ \textbf{\text{P}}_{\mathcal{B}}$ are point clouds for objects $\mathcal{A}$ and $\mathcal{B}$ respectively.  In other words, sampling from $f_D(\textbf{\text{P}}_{\mathcal{A}}, \textbf{\text{P}}_{\mathcal{B}})$ gives an SE(3) transformation $\textbf{T}_{\mathcal{A}\mathcal{B}}$ such that $\text{RelPlace}_D(\textbf{\text{T}}_{\mathcal{A}\mathcal{B}}~\cdot~\textbf{\text{T}}_{\mathcal{A}}, \textbf{\text{T}}_{\mathcal{B}}) =\textbf{SUCCESS}$. We refer to any such transformation $\textbf{T}_{\mathcal{A}\mathcal{B}}$ as a cross-pose, and the function $f_D(\textbf{\text{P}}_{\mathcal{A}}, \textbf{\text{P}}_{\mathcal{B}})$ outputs a distribution over cross-poses.

For this paper, we also assume that the demonstrations are given by segmented point clouds, in which we have a segmentation between the objects $\mathcal{A}$ and $\mathcal{B}$ in the demonstration.

\begin{figure*}[t]
    \centering
    \vspace{0.3cm} %
    \includegraphics[width=0.9\textwidth]{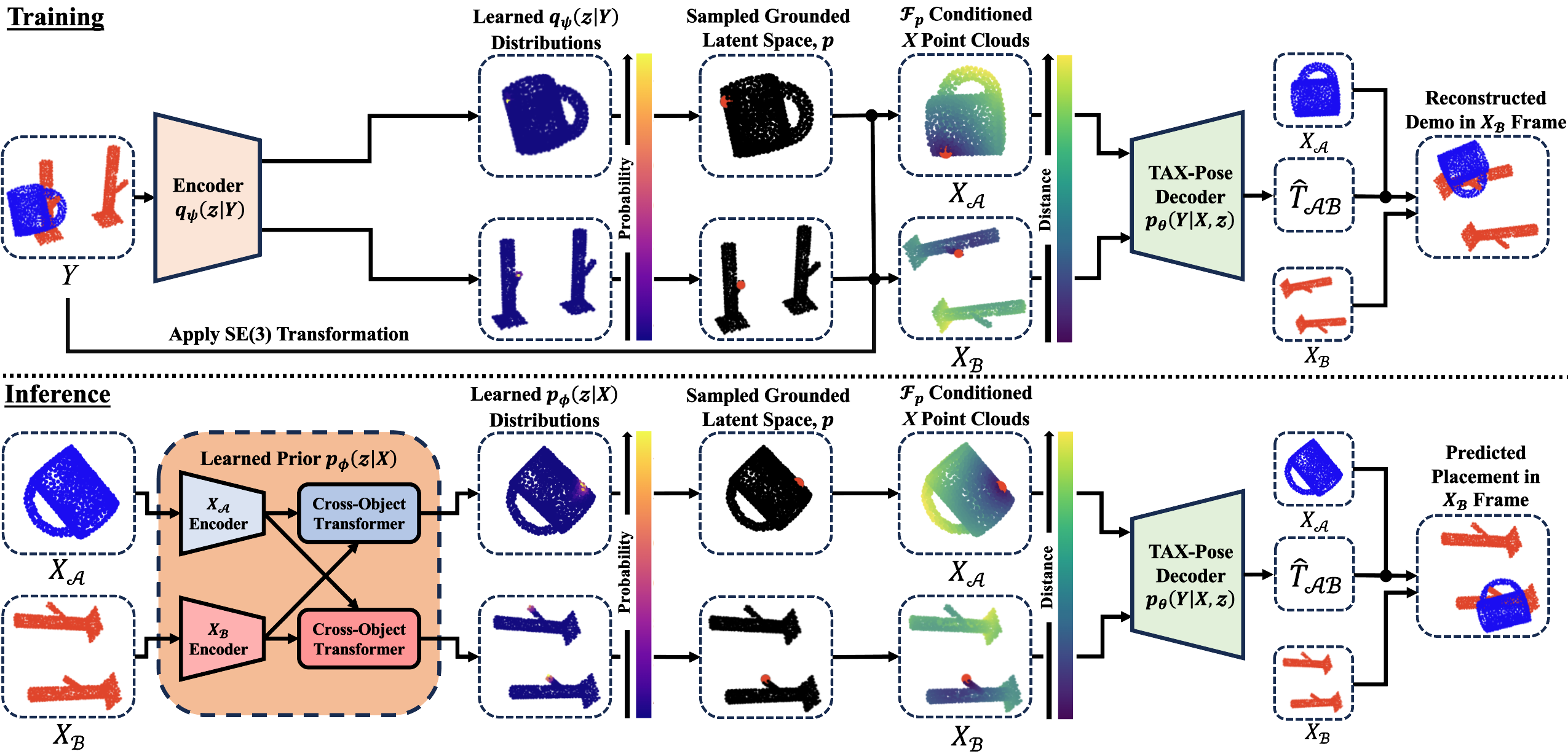}
    \caption{Model overview. We train an encoder $q_\psi(z \mid Y)$ to map from demonstration placements $Y$ into a spatially grounded latent variable $z$. Using a dense and SE(3)-invariant representation of $z$, $\mathcal{F}_p$, we condition the TAX-Pose decoder, $p_\theta(Y \mid X, z)$, to predict the cross-pose $\hat{\textbf{T}}_{\mathcal{A}\mathcal{B}}$ that reconstructs $Y$ from the SE(3) transformed action and anchor point clouds, $X_\mathcal{A}$ and $X_\mathcal{B}$. Additionally, we train a learned prior $p_\phi(z \mid X)$ to map from points $X$ to points in the latent space, capturing the different modes of the demonstrations. At inference time, we decode samples from this learned prior to generate valid cross-poses for the objects in the observation $X$. }
    \label{fig:method_overview}
    \vspace{-0.5cm}
\end{figure*}

\section{Method}
\label{sec:method}

We present TAX-PoseD, a method for relative placement tasks that accounts for multimodality over task solutions by sampling from a learned latent distribution over modes in the solution set.
Using a Conditional Variational Autoencoder (cVAE)~\cite{sohn2015learning}, our encoder maps a successful task demonstration to a distribution from which we sample a latent $z$, and our decoder maps $z$ to a reconstruction of the same scene. During inference time, we sample a latent $z$ from a prior and then decode it to a (hopefully) successful placement pose. However, we find that a traditional cVAE architecture of using a latent $z$ vector defined in an arbitrary vector space fails to learn  meaningful modes for this task, leading to poor performance.  In contrast,   
we propose a spatially-grounded latent space over points in the point cloud
which leads to significantly improved performance. Below, we describe our approach in more detail.

\subsection{Relative Placement as a Variational Problem}
\label{ssec:deriving_a_variational_loss}

Conditional Variational AutoEncoders (cVAEs)~\cite{sohn2015learning} are a variant of VAEs~\cite{kingma2013auto} that condition the encoder and decoder on additional inputs to control the data generation process. Given an input variable $X$ and output variable $Y$, the tractable variational lower bound of a cVAE (the ELBO) is written as the following~\cite{sohn2015learning}: 
\begin{align}
    \log p_\theta(Y \mid X) \geq \text{ }&\mathbb{E}_{q_\psi(z \mid X, Y)}\left[\log p_\theta(Y \mid X, z)\right] \label{eq:cvae} \\
    &- K L\left(q_\psi(z \mid X, Y) \| p_\phi(z \mid X)\right)  \nonumber
\end{align}
with the prior $p_\phi(z \mid X)$ and an encoder $q_\psi(z \mid X, Y)$ for latent variable $z$. The ELBO (Eq.~\ref{eq:cvae}) can be thought of as the sum of a reconstruction loss (first term) and a regularization loss (second term). 
In the relative placement setting, $Y$ is the point cloud of objects in a demonstration pose for a relative placement task, and $X$ is an observation of these objects in arbitrary poses.
During training, we obtain $X$ by transforming the demonstration objects $Y$ with arbitrary transforms sampled from SE(3).  Our goal is to learn a decoder $p_\theta(Y \mid X, z)$ that, conditioned on a latent $z$ and the observation $X$, predicts the object configuration in the demonstration $Y$.

As we note below, this straight-forward implementation of a cVAE does not work well for learning to imitate demonstrations of relative placement tasks.  
Below, we discuss the modifications we make to the cVAE framework for our task.
For example, rather than predicting a point cloud $Y$ directly, we instead %
predict a cross-pose $\textbf{T}_{\mathcal{A}\mathcal{B}}$ that can be applied to the objects in the observation $X$ to move them into the demonstration configuration. The robot can then move object $\mathcal{A}$ by  transform $\textbf{T}_{\mathcal{A}\mathcal{B}}$ to move it into the goal pose. Further, our experiments show that a continuous latent variable $z$ is not appropriate for relative placement tasks that have discrete multimodalities, such as placing a mug on one of $k$ distinct racks.  Thus, we instead use a discrete, spatially grounded latent space, as described below. %

\subsection{Learning a spatially-grounded latent space for demonstrations}
\label{ssec:method-step-1}

\textbf{Encoder:} First, we learn an encoder $q_\psi(z \mid X, Y)$ that compresses demonstration point clouds $Y$ into a latent space $z$. Since we focus on relative placement tasks, the demonstration point cloud $Y$ indicates an arrangement of objects that achieve a successful relative placement, and the latent encoding $z$ is a compressed form of this demonstration. 

As explained in Section~\ref{sec:problem_statement}, for relative placement tasks, an
object $\mathcal{A}$ must be placed in some configuration relative to another
object $\mathcal{B}$. The demonstration point clouds $Y$ show examples of successful relative placements of objects $\mathcal{A}$ and $\mathcal{B}$.
During training, we obtain the input $X$ by transforming the objects $\mathcal{A}$ and $\mathcal{B}$ in the demonstration point cloud $Y$ with arbitrary transformations sampled from SE(3). Thus, 
to encode the demonstration $Y$ we do not need knowledge of $X$.  We can thereby simplify the encoder $q_\psi(z \mid X, Y)$ as just $q_\psi(z \mid Y)$ since $z$ is conditionally independent of $X$ given $Y$, %
as shown in Figure~\ref{fig:method_overview}. More details about the encoder implementation can be found in Appendix~\ref{app:training-details} on the website.

\textbf{Decoder:} 
For the decoder $p_\theta(Y \mid X, z)$, we must 
decode the latent $z$ and the object point cloud $X$ into the demonstration point cloud $Y$.  %
However, training a decoder to predict the demonstration point cloud $Y$ directly would require the decoder to output unnecessarily complex details about the scene. Instead, we train the decoder to predict a transform $\textbf{T}_{\mathcal{A}\mathcal{B}} \in SE(3)$ that would move the 
objects in $X$ to their corresponding poses in the demonstration $Y$ (referred to as a ``cross-pose", see Section~\ref{sec:problem_statement} for details).  Further, this predicted cross-pose $\textbf{T}_{\mathcal{A}\mathcal{B}}$ is more useful than predicting a point cloud $Y$ since we can then command the robot to transform the objects in $X$ by transform $\textbf{T}_{\mathcal{A}\mathcal{B}}$ to achieve the task. 
To make this more clear, we will write the decoder below as $p_\theta(\textbf{T}_{\mathcal{A}\mathcal{B}} \mid X, z)$, to demonstrate that it outputs a distribution over transforms $\textbf{T}_{\mathcal{A}\mathcal{B}}$. 
In practice, the decoder will be implemented as a latent-conditioned version of TAX-Pose~\cite{pan2023tax}; 
more details about the decoder implementation can be found in Appendix~\ref{app:training-details} on the website.

In practice, during inference, we will first sample a latent $z$ from the prior $ p_\phi(z \mid X)$.  Conditioned on this sampled $z$ and the input $X$, we will then use the decoder $p_\theta(\textbf{T}_{\mathcal{A}\mathcal{B}} \mid~X, z)$ to output a (deterministic) transform $\hat{\textbf{T}}_{\mathcal{A}\mathcal{B}}$.  We will %
then command the robot to move object $\mathcal{A}$ by the sampled transform to achieve the relative placement task. Thus, combining the prior distribution with the deterministic decoder together gives a distribution over transforms $\textbf{T}_{\mathcal{A}\mathcal{B}}$.

\textbf{Reconstruction loss:}
As shown in Eqn.~\ref{eq:cvae}, the encoder and decoder are supervised using a reconstruction loss and a regularization loss.  In more detail, during training we first encode the demonstration point cloud $Y$ into a distribution over latent $z$ using the encoder $q_\psi(z \mid Y)$.  After sampling $z$ from this distribution, we decode the latent $z$ and the input $X$ using the decoder $p_\theta(\textbf{T}_{\mathcal{A}\mathcal{B}} \mid X, z)$ into a predicted transform $\hat{\textbf{T}}_{\mathcal{A}\mathcal{B}}$.
The reconstruction loss term $\mathbb{E}_{q_\psi(z \mid X, Y)}\left[\log p_\theta(\textbf{T}_{\mathcal{A}\mathcal{B}} \mid X, z)\right]$ corresponds to the difference between the demonstration point cloud $Y$ and the reconstructed scene after applying the predicted transform $\mathbf{T}_{\mathcal{A}\mathcal{B}}$ to the point cloud for object $\mathcal{A}$.  %

In more detail, let $X_\mathcal{A}$ and $X_\mathcal{B}$ be the point clouds for object $\mathcal{A}$ and $\mathcal{B}$ respectively.
Then let us define the reconstructed point cloud $\hat{Y} = (\hat{\mathbf{T}}_{\mathcal{A}\mathcal{B}} \cdot X_\mathcal{A}) \cup X_\mathcal{B}$ as the result of applying the predicted transform $\hat{\mathbf{T}}_{\mathcal{A}\mathcal{B}}$ to point cloud $X_\mathcal{A}$ and then concatenating point cloud $X_\mathcal{B}$. We will train the encoder and decoder such that, if the encoder $q_\psi(z \mid Y)$ and the decoder $p_\theta(\textbf{T}_{\mathcal{A}\mathcal{B}} \mid X, z)$ were lossless, then $\hat{Y}$ would be identical to the original demonstration point cloud $Y$.
The reconstruction loss computes the distance between the demonstration point cloud $Y$ and the reconstructed point cloud $\hat{Y}$, using similar losses as in TAXPose~\cite{pan2023tax}
(see Appendix~\ref{app:training-details} for details).

\textbf{Latent space:}
As mentioned above, we encode the demonstration point cloud $Y$ into a latent $z$ using the encoder $q_\psi(z \mid Y)$.
However, we found it difficult to train such a model if we represent the latent $z$ as an arbitrary vector in $\mathbb{R}^d$ like in traditional VAEs (see ablations in \autoref{ssec:ablations}).  We believe that this is because of the ``smoothing" effect of cVAEs; if~there are two discrete mug racks, some portion of the latent $z$ space should model placing the mug on rack 1, and some portion of the latent $z$ space should model placing the mug on rack 2.  However, because the latent space is smooth, there must be points that lie in the transition region between these spaces, and the model would have difficulty mapping such points to a valid output.  Further, if the next scene has 3 racks, then the latent space must be divided further, creating training difficulties with consistently mapping the latent space to the set of valid solutions.

\begin{figure}[t]
    \centering
    \vspace{0.3cm} %
    \includegraphics[width=0.48\textwidth]{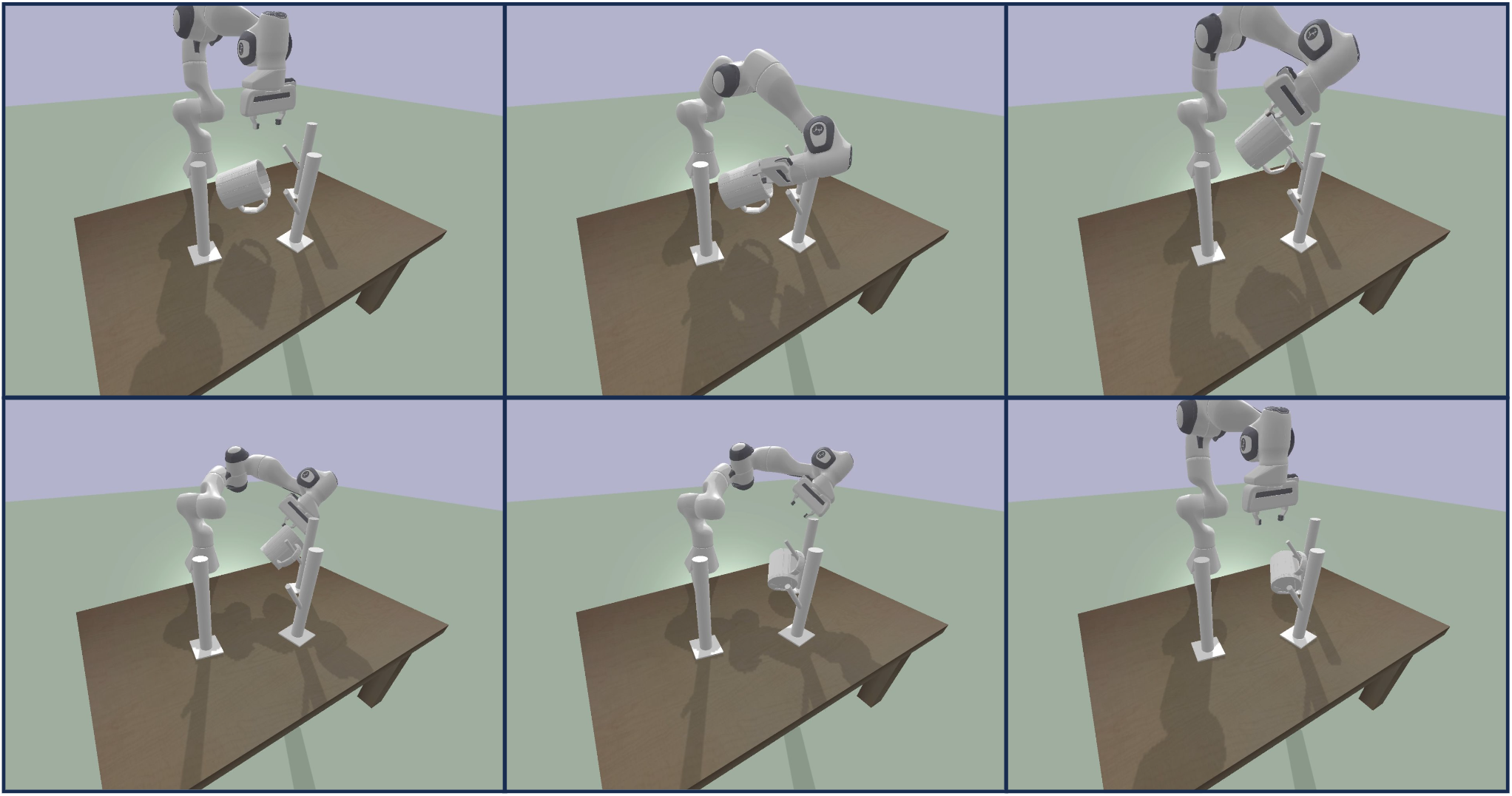}
    \caption{Visualization of a successful relative placement for hanging a mug with an arbitrary pose on one of many racks in a random 3 rack scene. The mug is initialized to an arbitrary pose in the air and must be grasped first by the robot.}
    \label{fig:successful_eval_place}
    \vspace{-0.5cm}
\end{figure}

We avoid these issues by training a \emph{spatially-grounded} latent space.
We represent the latent variable $z$ as a sample from a discrete $N$-dimensional distribution, where $N$ is the number of points in the point cloud $X$; in other words, the latent variable $z$ is represented as a categorical distribution over the points in the scene. Thus, the prior $p_\phi(z \mid X)$ as well as the encoder $q_\psi(z \mid Y)$ both represent multinomial distributions over all of the points in the point clouds $X$ and $Y$ respectively.
In practice, we actually represent the latent space by two separate categorical distributions, one over the points in $X_\mathcal{A}$ and another over the points in $X_\mathcal{B}$ (and similarly for the points in $Y$); we thus sample a latent point $p_\mathcal{A}$ among the points in object $\mathcal{A}$ and another latent point $p_\mathcal{B}$ among the points in object $\mathcal{B}$, and we condition the decoder on both sampled latent points (see the next section for details).  

The spatially-grounded latent distribution has a nice intuitive interpretation: the sampled latent points $z$ indicates which of the demonstration modes is selected.  For example, for the task of hanging a mug on one of $N$ racks, the latent point in $X_\mathcal{B}$  empirically tends to indicate which of the racks will be used for hanging (see a visualization of this latent space in \autoref{fig:p(z|X)_vis.jpg}. %
Thus, this spatially grounded latent space is able to interpretably select a mode from the demonstration distribution by conditioning on  a learned distribution over points from $X_\mathcal{A}$ and  $X_\mathcal{B}$.  Because this distribution is discrete, we avoid the ``smoothing" issue normally found with cVAEs mentioned in the previous paragraph.
In order to back-propagate through these discrete latent distributions, we perform  differentiable sampling %
with Gumbel Softmax~\cite{jang2016categorical}. %

Architecturally, we design the prior $p_\phi(z \mid X)$ as well as the encoder $q_\psi(z \mid Y)$ to predict a value over each point in the point clouds $X$ and $Y$ respectively, as shown in Figure~\ref{fig:method_overview}.  We then normalize these values to obtain a discrete probability distribution over the points in the point clouds.

\textbf{Latent conditioning:}
As mentioned previously, we
use a modified version of TAX-Pose~\cite{pan2023tax} for the decoder $p_\theta(\textbf{T}_{\mathcal{A}\mathcal{B}}~\mid~X, z)$.
However, we must modify the TAX-Pose model to condition on the latent points $z = (p_\mathcal{A}, p_\mathcal{B})$. 
One option for doing so would be to represent $z$ as the discrete value selected from the $N$-dimensional multinomial distribution over points.  However, the discrete value would correspond to the index of the selected point, which  has no particular meaning to the network, which uses a permutation-invariant point cloud-based architecture~\cite{wang2019dynamic}.  %

Another  option is to use the 3D location of the points  $(p_\mathcal{A}, p_\mathcal{B})$  directly; however, we would like the model to be translation-equivariant, so we would like to  avoid explicitly inputting the 3D locations of the latent points.  Another alternative is to concatenate to each input a binary indicator of whether that point was selected as the latent value $z$.  However, such a sparse input is very difficult for the decoder to process.  

Instead, we create a dense and SE(3)-invariant representation of $z$, as follows: For each point $p \in X_\mathcal{A}$, we compute a feature $\mathcal{F}_p \defeq  \|p - p_\mathcal{A} \| _2$, and likewise for $X_\mathcal{B}$, which indicates the distance between each point $p$ and the latent point $p_\mathcal{A}$ or $p_\mathcal{B}$.  This feature is both translationally- and rotationally-invariant, in that if we translate or rotate the entire point cloud $X_\mathcal{A}$, then $\mathcal{F}_p$ will remain the same (see Appendix~\ref{app:rotationally-invariant-features} for the proof).  We concatenate $\mathcal{F}_p$ to the input of each point in $X$ in order to condition the decoder on the latent variable $z$. A~visualization of $\mathcal{F}_p$ is depicted in Figure~\ref{fig:method_overview}.

The TAX-Pose~\cite{pan2023tax} model that we use for the decoder achieves translation invariance by mean-centering the points in $X_\mathcal{A}$ and $X_\mathcal{B}$.  However, for larger multi-object scenes such as those with multiple racks explored in this work, mean-centering doesn't lead to a consistent point cloud observation.  To address this, for the decoder $p_\theta(\textbf{T}_{\mathcal{A}\mathcal{B}} \mid X, z)$, we 
center the points in $X_\mathcal{A}$ (and $X_\mathcal{B}$ respectively) on the sampled points $p_\mathcal{A}$ and $p_\mathcal{B}$. %
Empirically, we find that these architectural choices lead to significantly improved performance %
for relational object placement tasks.

\subsection{Generalizing to arbitrary configurations for inference}
\label{ssec:method-step-2}

During  cVAE training, we also include a regularization loss $K L\left(q_\psi(z \mid X, Y) \| p_\phi(z \mid X)\right)$ %
as shown in the second term of Equation~\ref{eq:cvae}. This regularization term  allows us to sample from the prior $p_\phi(z \mid X)$ at inference time and generate a latent $z$ that is similar to the distribution of encoded latent z values $q_\psi(z \mid X, Y)$.
Normally, the prior distribution $p_\phi(z \mid X)$ is set to be a Normal Gaussian distribution; in our case, the latent space $z$ is discrete, so we could set the prior to a uniform distribution.

However, a uniform prior would not meaningfully encode the modes of the demonstration space.  As shown in \autoref{fig:p(z|X)_vis.jpg}, the prior $p_\phi(z \mid X)$ should map the input to a distribution that encodes the modes of the demonstrations.  For example, in a scene with multiple mug racks, the distribution over $z$ would encode the different options of selecting a mug rack.  A~uniform prior $z$ over all points in the scene would encourage an equal probability mass on points that are not mug racks at all, and thus the latent $z$ would not meaningfully encode the demonstration modes. We observe that using a uniform prior leads to poor generalization in practice.  Thus, we train the encoder and decoder %
without an explicit regularization loss. %

Instead, we train a learned prior $p_\phi(z \mid X)$ to map from the observation $X$ to points in the latent space over $z$ that encode the different modes of the demonstrations.  We train the prior using only the regularization loss from Equation~\ref{eq:cvae}:
\begin{align}
    \mathcal{L} \defeq K L\left(q_\psi(z \mid X, Y) \| p_\phi(z \mid X)\right)
\end{align}
We apply a stop gradient on this loss so that we do not back-propagate through the encoder $q_\psi(z \mid X, Y)$, which 
we separately (and simultaneously) %
 train using the reconstruction loss to map from a demonstration pose $Y$ to a point in the latent space $z$ to capture each demonstration mode.
Because the learned prior $p_\phi(z \mid X)$ does not condition over the demonstration $Y$, it must marginalize over all demonstrations in the dataset.  Thus, for the example of hanging a mug with multiple racks, the encoder
$q_\psi(z \mid X, Y)$ will map from a demonstration $Y$ to a distribution over the points on the rack that the mug is hung on in the demonstration $Y$, whereas the learned prior $p_\phi(z \mid X)$ will map from the input $X$ to a distribution over \emph{all} of the mug racks as shown in \autoref{fig:p(z|X)_vis.jpg}.
Further implementation details on the learned prior can be found in Appendix~\ref{app:training-details} on the website.

The inference procedure for our method can be found in \autoref{fig:method_overview} (bottom).  We first input the observation $X$ into a learned prior $p_\phi(z \mid X)$ which outputs a distribution over the latent space $z$.  We sample a $z$ from this distribution, which we convert to a dense rotationally-invariant feature $\mathcal{F}_p$ as described above.  Finally, we input $X$ and our representation of $z$ into the modified TAX-Pose~\cite{pan2023tax} decoder $p_\theta(\textbf{T}_{\mathcal{A}\mathcal{B}} \mid X, z)$ %
to output the predicted cross-pose $\hat{\textbf{T}}_{\mathcal{A}\mathcal{B}}$.  We then send this predicted cross-pose to our robot system, which uses motion planning~\cite{sucan2012the-open-motion-planning-library} to move the action object $\mathcal{A}$ such that it is transformed by transformation $\hat{\textbf{T}}_{\mathcal{A}\mathcal{B}}$. If the learned prior and decoder are trained well, then this should complete the relative placement task.

\section{Experiments}
\label{sec:experiments}

\subsection{Environments}

We evaluate our method on the NDF~\cite{simeonov2022neural} simulated mug placement tasks in addition to multi-modal variants of the original mug-hanging task. In all these tasks, the goal for the robot is to move an action object to a pose relative to the rest of the scene. The robot is given only 10 demonstrations, from which we extract a point cloud recording of the final configuration from each demonstration. The tasks are implemented in Pybullet~\cite{coumans2021}, in which a simulated robot is placed on a table, surrounded by four depth cameras. We increase the difficulty of the original mug-hanging NDF task to require reasoning under multimodality: instead of one rack, the environment contains multiple racks in various configurations. %

We implement our method for each relative placement task with two sequential cross-pose estimation steps: the cross-pose between the gripper and the action object (e.g. the mug) for grasping, followed by the cross-pose between the object and the rest of the scene (e.g. the mug racks) for placing. %
We use motion planning to move the gripper into the desired positions for both parts of the task; after  the gripper rigidly grasps the mug, the  mug-on-rack cross-pose defines the desired target pose of the gripper. 
In all of our evaluations, the object (e.g.~mug)  is initialized to an arbitrary initial pose in space. 
A placement is marked as a success when the object is placed stably on one of the target placement locations.

\begin{table}[t]
    \vspace{0.3cm} %
    \sisetup{round-mode=places,round-precision=2} %
    \centering %
    \begin{tabular}{@{} l  cc|c @{}}
    \toprule
    &  &  & \textbf{(Generalization)} \\
    & \textbf{1 Rack Place} & \textbf{2 Racks Place} & \textbf{3 Racks Place} \\
        
    \midrule
    DON~\cite{florence2018dense} & 0.45 & - & - \\
    NDF~\cite{simeonov2022neural} & 0.75 & - & -  \\
    TAX-Pose~\cite{pan2023tax} & \textbf{0.84} & 0.07 & 0.14 \\
    \textbf{TAX-PoseD (Ours)} & 0.82 & \textbf{0.57} & \textbf{0.59} \\
    \bottomrule
    \end{tabular}
    \caption{\textbf{Task success rates for the original NDF 1 rack mug hanging task and multimodal variants.} We evaluate on unseen mugs initialized to arbitrary configurations in space. \textit{1~Rack:} Our model performs on par with TAX-Pose~\cite{pan2023tax} in the unimodal placement task and outperforms the other baselines. \textit{2~Racks:} We extend TAXPose to a multimodal task and train our model for random 2-rack scenes. We evaluate on 50 random 2-rack configurations. \textit{3~Racks (Generalization):} We further evaluate the 2 rack model to test generalization to more racks than seen at training time. Three racks are placed in 50 random configurations. We report placement task success.}
    \label{tab:original-ndf-tasks}
\end{table}

\subsection{Baseline Comparison}

We present task success evaluations for a variety of tasks with object instances and configurations that are unseen during training. We first evaluate our model on the ``1~Rack" mug hanging task from NDF~\cite{simeonov2022neural}, which requires no multimodality. 
In ``1 Rack Place", our model performs on par with TAX-Pose~\cite{pan2023tax} (\autoref{tab:original-ndf-tasks}) and outperforms other baselines in placement. Details of DON~\cite{florence2018dense} and NDF~\cite{simeonov2022neural} can be found in prior work~\cite{simeonov2022neural}.  %

Next, we %
show how our formulation extends TAX-Pose~\cite{pan2023tax} to multimodal mug-hanging tasks of increasing difficulty in \autoref{tab:original-ndf-tasks}.
We train and evaluate the TAX-Pose model and our method on the ``2 Racks" task,
in which the task is to hang the mug on either of two racks that each have an arbitrary pose on a tabletop.
We see a large improvement over the original TAX-Pose method in the 2-rack case in \autoref{tab:original-ndf-tasks}. We also test our model's ability to generalize to more racks %
than the number seen during training: despite only seeing two racks during training, our model is able to retain strong performance for the ``3 Racks" task. 
We observe that the TAX-Pose baseline tends to incorrectly predict mug placements in between the racks. %
Visually, we see that the learned prior $p_\phi(z \mid X)$ distribution will cover all of the racks available during inference, as shown in \autoref{fig:p(z|X)_vis.jpg}.

\subsection{Ablations}
\label{ssec:ablations}
In prior work, VAE variants often represent the latent $z$ as a global vector~\cite{higgins2016beta} sampled from a Normal distribution. %
Our method, in contrast, represents $z$ as a discrete distribution over points in the point cloud, which we refer to below as  ``spatial $z$" in \autoref{tab:ablations}. %
We provide ablations to our method for which we replace the  encoder $q_\psi(z \mid X, Y)$ with one that outputs a continuous $z$ vector like in prior work (``continuous $z$"). We condition the decoder with such a $z$ by following the architecture of the goal-conditioned TAX-Pose-GC~\cite{pan2023tax}. We also provide ablations for regularizing the encoder $q_\psi(z~\mid~X,~Y)$ to a Normal or Uniform prior. %
In \autoref{tab:ablations}, we show that the continuous $z$ ablations fail to predict precise placements for both multimodal environments, and the Uniform prior ablation for discrete spatial $z$ struggles to generalize to the ``3 Racks" environment. In contrast, our method is able to account for this multimodality and generalization.

    \setlength{\tabcolsep}{2.2pt}
    \begin{table}[t]
    \vspace{0.3cm}
    
    \sisetup{round-mode=places,round-precision=3} %
    \centering %
    \begin{tabular}{@{} l c c c | c  c c @{}}
    \toprule
    & \multicolumn{3}{c }{\textbf{2 Racks}} %
    & \multicolumn{3}{c}{\textbf{3 Racks}} \\
    \cmidrule(lr){2-4} \cmidrule(lr){5-7}
   & \textbf{Place} $\uparrow$ & \textbf{$\epsilon_{R}$} $\downarrow$ & \textbf{$\epsilon_{t}$} $\downarrow$ 
    & \textbf{Place} $\uparrow$ & \textbf{$\epsilon_{R}$} $\downarrow$ & \textbf{$\epsilon_{t}$} $\downarrow$
    \\
    \hline
    \rule{0pt}{10pt} %
    TAX-Pose~\cite{pan2023tax} & 0.07  &  103  & 0.45 & 0.14 & 71.8 & 0.33 \\
    \hline
    \rule{0pt}{10pt} %
    TAX-PoseD ablations: & & & &  \\
    \hspace{0.05cm} - continuous $z$, Normal prior & 0.33 &  70.2  & 0.28 & 0.18 & 66.2 & 0.31 \\
    \hspace{0.05cm} - continuous $z$, learned prior & 0.41 & 43.3 & 0.24 & 0.29 & 44.5 & 0.29 \\
    \hspace{0.05cm} - spatial $z$, uniform prior & \textbf{0.60}  &  14.1   & \textbf{0.05} & 0.18 & 25.3 & 0.25 \\
    \hspace{0.05cm} - \textbf{Ours: spatial $z$, learned prior} & 0.57 & \textbf{13.2} & \textbf{0.05} & \textbf{0.59} & \textbf{16.0} & \textbf{0.06} \\
    \bottomrule
    \end{tabular}
    \caption{\textbf{Ablations.} We show ablations of our model for the multimodal 2-racks mug-hanging task and also for generalization to the unseen 3-racks task. We report placement success, and also rotation error ($^{\circ}$) and translation error (m) relative to the closest demonstration. Using a discrete spatial $z$ distribution over points in the point cloud performs better than using a continuous z vector;  using a learned prior performs better than using a non-learned prior (Normal or uniform), especially for generalization to ``3 Racks."} %
    \label{tab:ablations}
    \vspace{-0.5cm}
\end{table}

\section{Conclusion}
\label{sec:conclusion}

We presented TAX-PoseD, a data-driven model for relative placement tasks that accounts for multimodality in task specification by learning a latent distribution over points in the object point clouds. %
We show that our method efficiently learns distributional relative placement tasks with only 10 demonstrations. %
Our method extends the VAE literature to practically implement an interpretable, spatially-grounded latent space. We achieve high task success on the mug placing task with a varying number of racks, %
which requires multimodal reasoning, with the use of SE(3)-invariant modules for reasoning about multimodal relative placement.

\section*{ACKNOWLEDGMENT}
Special thanks to the guidance provided by the TAX-Pose authors Chuer Pan, Brian Okorn, Harry Zhang, and Ben Eisner. %

\bibliographystyle{IEEEtran}
\bibliography{main_bib}

\clearpage

\appendix

\subsection{Training Details}
\label{app:training-details}

\textbf{Loss functions:}
We train the encoder and decoder with a reconstruction loss; the decoder outputs a transformation $\hat{\textbf{T}}_{\mathcal{A}\mathcal{B}}$ which we apply to the action object point cloud $X_\mathcal{A}$ to obtain a predicted point cloud $\hat{Y}$; we then compare this predicted point cloud to the demonstration ground-truth point cloud $Y$ to obtain the loss. We train these networks with similar  losses as were used in TAX-Pose~\cite{pan2023tax} losses, i.e.
a weighted sum of a Point Displacement Loss (weight=1), a Direct Correspondence Loss (weight=0.1), and a Correspondence Consistency Loss (weight=1). 

We also train the learned prior, $p_\phi(z \mid X)$, to match the distribution predicted by the encoder, $q_\psi(z \mid X, Y)$. In practice, while this can be implemented as the KL loss between the encoders' outputs $p_\phi(z \mid X)$ and $q_\psi(z \mid X, Y)$, empirically we find using the Jenson-Shannon Divergence (JSD) \cite{lin1991divergence} to improve the learned prior performance. For our spatially grounded latent space defined as a categorical distribution over the points in the scene, the JSD is defined as:

\begin{align}
    JSD(q_\psi, p_\phi) = & \text{ } \frac{1}{2}K L\left(q_\psi \| \frac{q_\psi + p_\phi}{2} \right) \\
    &+ \frac{1}{2}K L\left(p_\phi \| \frac{q_\psi + p_\phi}{2} \right) \nonumber
\end{align}
Because our network outputs separate distributions over the action object points $X_\mathcal{A}$ and the anchor object points $X_\mathcal{B}$, we take the sum of the JSD over the action object distribution and the JSD over the anchor objectdistribution. We use a learning rate of $\alpha =$\texttt{1e-4} and gradient clipping of $\texttt{1e-3}$ for all models.

\textbf{Network Architecture for Encoder:}
The encoder $q_\psi(z \mid Y)$ is implemented as a single-scale PointNet++ \cite{qi2017pointnet++} directly encoding the points of the action and anchor objects in $Y$ into per-point features, which are then normalized into probability distributions with a softmax. In more detail: we separately normalize the action and anchor object points to create two distributions: $q_\psi(z_\mathcal{A} \mid Y)$ and $q_\psi(z_\mathcal{B} \mid Y)$, from which we sample the spatially-grounded latent points $p_\mathcal{A}$ and $p_\mathcal{B}$. We use a single-scale PointNet++ for encoding demonstrations with $q_\psi(z \mid Y)$ to capture the local relationship between the action object and the single demonstration mode in the anchor object of $Y$. 

\textbf{Network Architecture for Learned Prior:}
The learned prior $p_\phi(z \mid X)$ is implemented as two DGCNN \cite{wang2019dynamic} networks that each separately encode the action and anchor point clouds, followed by a Transformer \cite{10.5555/3295222.3295349} that performs a cross-object comparison across the action and anchor objects.  The output of the transformer are per-point features, which are then each passed through a point-wise MLP and normalized into probability distributions (separately for the action and anchor points) with a softmax. %

\textbf{Network Architecture for Decoder:}
As previously described, we use a modified version of TAX-Pose as the decoder $p_\theta(\textbf{T}_{\mathcal{A}\mathcal{B}} \mid X, z)$. %
We modify TAXPose to condition on the latent $z$ as described in Sec.\ref{ssec:method-step-1}. %
Another minor modification that we make is as follows: within TAX-Pose's cross-correspondence estimator, two point-wise MLPs are used to map cross-object point embeddings into correspondence residuals and per-point weights. Empirically, we find that using a single point-wise MLP to jointly predict the correspondence residuals and per-point weights improves the performance of our method.

\textbf{Dataset generation:}
To generalize to multiple racks in our mug-hanging experiments, we create a training set with multi-rack demonstrations.  To do so, we modify the environment used for %
1-rack mug hanging to generate multi-rack demonstrations. %
We  create a distractor rack  by copying the points from the rack in the 1-rack demonstration with a random SE(3) transformation applied; we then  concatenate this distractor rack to the scene point cloud. We ensure that the scene is physically-plausible by performing axis-aligned rectangular prism collision checking between the original demonstration and the distractor rack. %
When training the encoder $q_\psi(z \mid Y)$, we do not apply any transformations to the points in the demonstration point cloud $Y$.
However, when we train the learned prior $p_\phi(z \mid X)$, we rotate both the original rack and the distractor rack to enable the learned prior to be robust to new rack poses that might be encountered at test-time.

To generate an arbitrary initial configuration $X$ from a demonstration $Y$, we uniformly sample two random SE(3) transformations and apply them to the points corresponding to objects $\mathcal{A}$ and $\mathcal{B}$, respectively, to produce a point cloud for each object in an arbitrary pose: $X_{\mathcal{A}}$ and $X_{\mathcal{B}}$. These two point clouds together form the point cloud $X = X_{\mathcal{A}} \cup X_{\mathcal{B}}$. We found it helpful to center the $X_{\mathcal{A}}$ and $X_{\mathcal{B}}$ input to the decoder by the sampled point from the latent distribution $p_{\mathcal{A}} \sim p(z_{\mathcal{A}} \mid X_{\mathcal{A}})$ and $p_{\mathcal{B}} \sim p(z_{\mathcal{B}} \mid X_{\mathcal{B}})$, respectively. The inputs to the encoders $q_\psi(z \mid X, Y)$ and $p(z \mid Y)$ are mean centered.

\textbf{Synthetic occlusions:}
We apply planar and ball synthetic occlusions to the action object as data augmentations simulating partially-occluded scenes. Specifically, the planar synthetic occlusion chooses a random 3D plane to be between a point in the point cloud and the point cloud center, then removes all points on one side of the plane. Also, the ball synthetic occlusion chooses a random point in the point cloud then removes all points within a certain radius of that chosen point. These occlusions can roughly simulate occlusions from a limited number of camera perspectives and various object interactions. In addition, when downsampling the demonstration point cloud, we apply a random downsampling 50\% of the time and apply furthest point downsampling 50\% of the time to capture more randomness in the distribution of points. %

\textbf{Motion Planning:}
The model's cross-pose predictions guide an OMPL motion planner for the robot to produce the executed trajectory, which takes into account obstacles and joint limits.

\subsection{Task Details}
Details about the different environments that we use for evaluation can be found in \autoref{tab:task-details}.

\begin{table}[h!]
    
    \vspace{0.3cm} %
    \sisetup{round-mode=places,round-precision=2} %
    \centering %
    \begin{tabular}{@{} p{2.5cm} p{5cm} @{}}
    \toprule
    \textbf{Environment} & \textbf{Description} \\
    \midrule
    Mug-on-Rack (1 rack) / 1 Rack (fixed) & Original NDF task of hanging a mug on a mug rack. The rack is at a fixed location on the table and there are a variety of unseen mugs at test time. \\
    2 Racks & The two mug racks are randomly rotated with respect to the z axis and randomly translated across the table surface. We evaluate on a fixed test set of 50 rack configurations. \\
    3 Racks & The three mug racks are randomly rotated with respect tot he z axis and randomly translated across the table surface. We evaluate on a fixed test set of 50 rack configurations. \\
    \bottomrule
    \end{tabular}
    \caption{Environment Descriptions}
    \label{tab:task-details}
\end{table}
\FloatBarrier

\subsection{Proof for rotational invariance for the featurization}
\label{app:rotationally-invariant-features}

Here, we prove that our latent conditioning described in~\autoref{ssec:method-step-1} is SE(3)-invariant.

Let there be an arbitrary transformation $T \in SE(3)$ that rigidly transforms an object in an arbitrary pose $X$ to the same object in a demonstration pose $Y$. Transformation $T$ can be rewritten as a rotation $R$ and translation component $t$.

$$X \defeq T Y$$

Let $p_x$ be a point in point cloud $X$ where the feature is being computed, and let $z_x$ be the point sampled from the latent distribution. Define $p_y$ and $z_y$ similarly for point cloud $Y$. Then, our method's featurization is invariant to SE(3) transformations with respect to the demonstration $Y$:

\begin{align}
    \mathcal{F} \defeq ||p_x - z_x||_2 &= ||T^{-1}p_y - T^{-1}z_y||_2 \nonumber \\ &= ||R^T(p_y - t - z_y + t)||_2 \nonumber \\ &= ||p_y - z_y||_2
\end{align}

Note that without the L2 norm, the featurization would not be rotationally-invariant.

\end{document}